\documentclass[sigconf]{acmart}

\copyrightyear{2023}
\acmYear{2023}
\setcopyright{acmlicensed}\acmConference[MM '23]{Proceedings of the 31st ACM International Conference on Multimedia}{October 29-November 3, 2023}{Ottawa, ON, Canada}
\acmBooktitle{Proceedings of the 31st ACM International Conference on Multimedia (MM '23), October 29-November 3, 2023, Ottawa, ON, Canada}
\acmDOI{10.1145/3581783.3613769}
\renewcommand\footnotetextcopyrightpermission[1]{}
\usepackage{gensymb}
\usepackage{amsmath}
\usepackage{hyperref}
\usepackage{cleveref}
\usepackage{bm}
\usepackage{caption}
\usepackage{graphicx}
\usepackage{float}
\usepackage{booktabs}
\usepackage{multirow}
\usepackage{subfigure}
\usepackage{enumitem}
\crefname{equation}{Eq.}{Eqs.}
\crefname{figure}{Fig.}{Figs.}
\crefname{section}{Sec.}{Secs.}
\usepackage{balance}
\begin{document}

\title{Where and How: Mitigating Confusion in Neural Radiance Fields from Sparse Inputs}

\author{Yanqi Bao}
\email{yq_bao@smail.nju.edu.cn}
\orcid{0000-0001-5298-7087}
\affiliation{%
  \institution{Nanjing University}
  \city{Nanjing}
  \state{Jiangsu}
  \country{China}
}

\author{Yuxin Li}
\email{liyuxin16@smail.nju.edu.cn}
\affiliation{%
  \institution{Nanjing University}
  \city{Nanjing}
  \state{Jiangsu}
  \country{China}
}

\author{Jing Huo}
\email{huojing@nju.edu.cn}
\affiliation{%
  \institution{Nanjing University}
  \city{Nanjing}
  \state{Jiangsu}
  \country{China}
}

\author{Tianyu Ding}
\email{tianyuding@microsoft.com}
\affiliation{%
  \institution{Microsoft Corporation}
  \state{Redmond}
  \country{USA}
}

\author{Xinyue Liang}
\email{mf21330051@smail.nju.edu.cn}
\affiliation{%
  \institution{Nanjing University}
  \city{Nanjing}
  \state{Jiangsu}
  \country{China}
}

\author{Wenbin Li}
\email{liwenbin@nju.edu.cn}
\affiliation{%
  \institution{Nanjing University}
  \city{Nanjing}
  \state{Jiangsu}
  \country{China}
}

\author{Yang Gao}
\email{gaoy@nju.edu.cn}
\affiliation{%
  \institution{Nanjing University}
  \city{Nanjing}
  \state{Jiangsu}
  \country{China}
}

\renewcommand{\shortauthors}{Yanqi Bao et al.}

\begin{abstract}
  \textit{Neural Radiance Fields from Sparse inputs} (NeRF-S) have shown great potential in synthesizing novel views with a limited number of observed viewpoints. However, due to the inherent limitations of sparse inputs and the gap between non-adjacent views, rendering results often suffer from over-fitting and foggy surfaces, a phenomenon we refer to as "CONFUSION" during volume rendering. In this paper, we analyze the root cause of this confusion and attribute it to two fundamental questions: "WHERE" and "HOW". To this end, we present a novel learning framework, WaH-NeRF, which effectively mitigates confusion by tackling the following challenges: (i) \textbf{“WHERE” to Sample?} in NeRF-S---we introduce a Deformable Sampling strategy and a Weight-based Mutual Information Loss to address sample-position confusion arising from the limited number of viewpoints; and (ii) \textbf{“HOW” to Predict?} in NeRF-S---we propose a Semi-Supervised NeRF learning Paradigm based on pose perturbation and a Pixel-Patch Correspondence Loss to alleviate prediction confusion caused by the disparity between training and testing viewpoints. By integrating our proposed modules and loss functions, WaH-NeRF outperforms previous methods under the NeRF-S setting. Code is available https://github.com/bbbbby-99/WaH-NeRF.
\end{abstract}



\keywords{Neural Radiance Field from Sparse Inputs; Volume Rendering; Semi-Supervised Learning}

\maketitle
\section{Introduction}\label{Introduction}
The field of 3D modeling and scene synthesis has experienced a significant transformation with the advent of Neural Radiance Fields (NeRF)~\cite{mildenhall2021nerf}. By utilizing deep neural networks, NeRF and its variants feature implicit representations of entire 3D scenes, resulting in remarkable progress in generating geometric 3D representations and synthesizing immersive novel views. Unlike traditional Novel View Synthesis methods ~\cite{sajjadi2022scene,flynn2019deepview}, NeRF exhibits higher efficiency through volume rendering. However, NeRF's ability to generate realistic renderings relies heavily on dense viewing inputs, which can be challenging to collect in real-world applications such as robotic navigation~\cite{rosinol2022nerf} and autonomous driving ~\cite{tancik2022block}, among others.

In scenarios with insufficient inputs, several works ~\cite{xu2022sinnerf,kim2022infonerf} have demonstrated the negative impact on vanilla NeRF performance and introduced \textit{Neural Radiance Fields from Sparse inputs} (NeRF-S) to address this challenge. Unfortunately, sparse viewpoints often fail to provide enough information to generate a complete geometric 3D representation, leading to some derivative works of NeRF ~\cite{yu2021pixelnerf,wang2021ibrnet} as well as traditional NVS methods focusing on adjacent viewpoints. However, this simplification restricts the rendering scope and application scenarios for NeRF, prompting recent research~\cite{niemeyer2022regnerf, kim2022infonerf} efforts to explore rendering a complete scene under sparse inputs.

\begin{figure}
\centering 
\includegraphics[width=1\linewidth,scale=0.6]{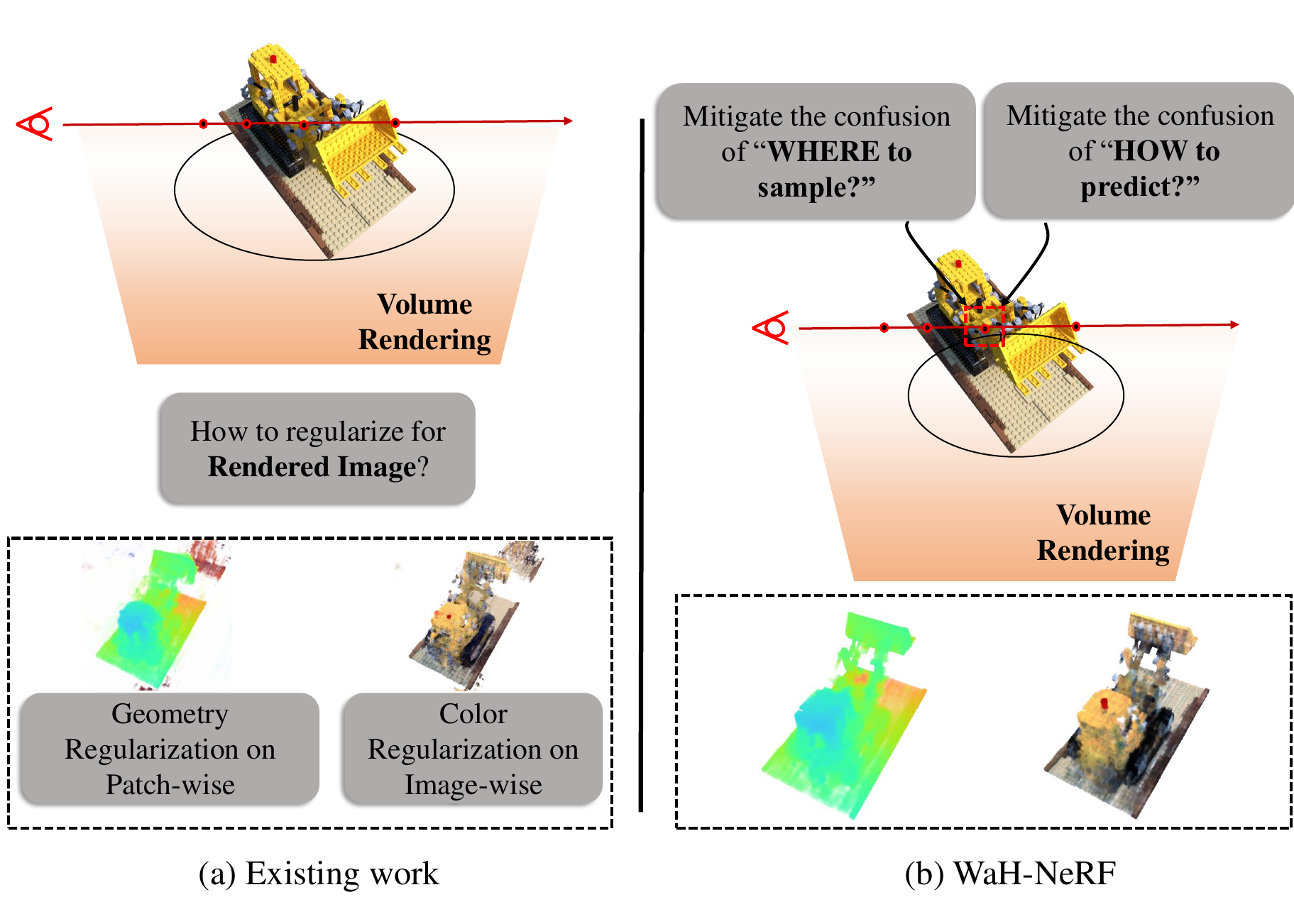}
\caption{\textit{Different from existing algorithms}: Existing state-of-the-art methods, e.g. RegNeRF~\cite{niemeyer2022regnerf} as (a), rely on regularization for rendered color and depth images, which ignore the causes of performance degradation before volume rendering. Our work, as shown in (b), analyzes the sample-render process and mitigates the confusion of “WHERE to sample?” and “HOW to predict?” in NeRF-S during volume rendering.}
\label{figure_1}
\end{figure}

For this, recent prominent works ~\cite{niemeyer2022regnerf, xu2022sinnerf} propose image-wise (or patch-wise) regularization (as~\Cref{figure_1}(a)) on geometry and color by adding additional supervision ~\cite{niemeyer2022regnerf} or warping for pseudo-label ~\cite{xu2022sinnerf}. Although these approaches alleviate the gap between training and testing viewpoints and achieve state-of-the-art performance, the additional supervision, such as depth image and pre-trained models, are often not feasible, and the warping can be time-consuming. 

\begin{figure*}
\centering
\includegraphics[width=\linewidth,scale=0.3]{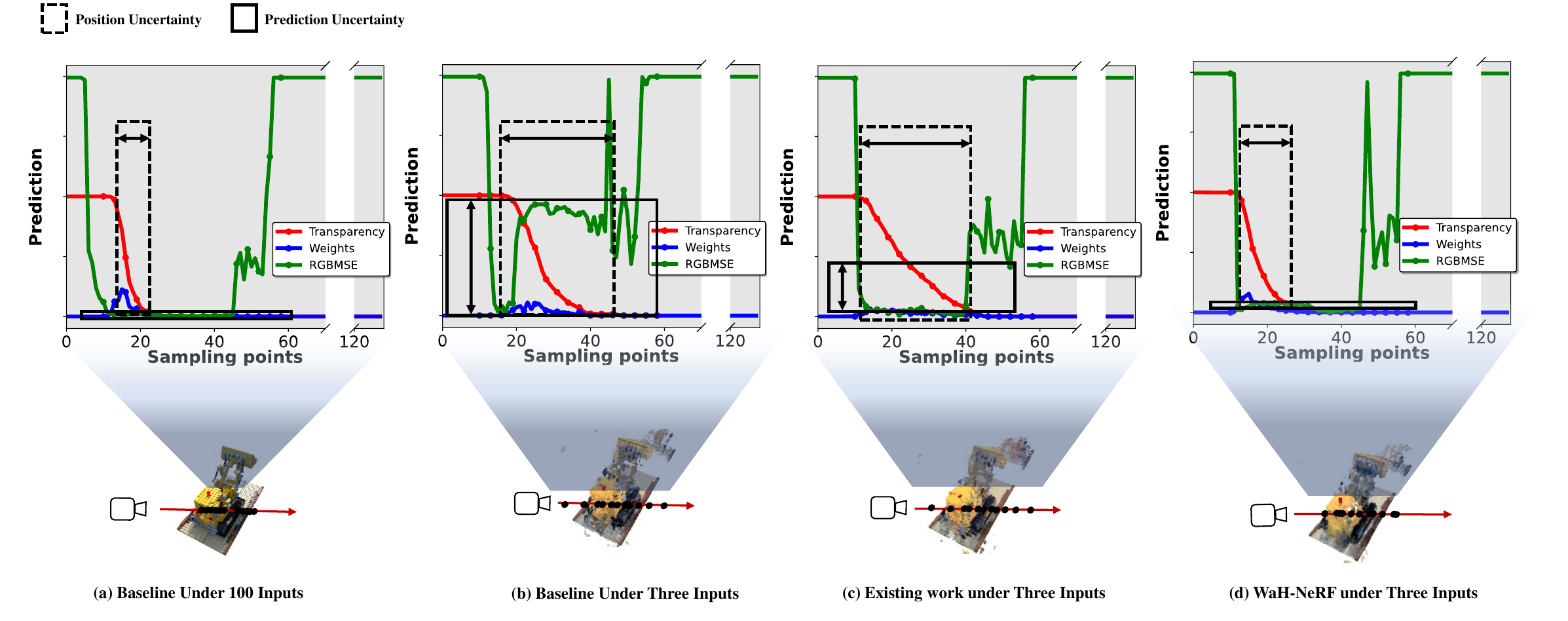}
\caption{\textit{Uncertainty analysis under sparse inputs.} We perform an analysis on the 128 sampling predictions of a random ray, which includes the RGBMSE, Weight and Transparency. RGBMSE is the mean squared error loss between the RGB color of each sampling position and the ground truth at corresponding pixel, representing the uncertainty of the color prediction. The Weight and Transparency are intermediate outputs of the network, which approximate the position of the surface. We use the width of the rectangular box to represent the uncertainty (or confusion) about position and prediction. Our results show that WaH-NeRF successfully mitigates the confusion caused by sparse inputs, which is further elaborated in the Appendix.} 
\label{figure_12}
\end{figure*}

In this work, instead of relying on warping or additional supervision on rendered images, we approach this problem by analyzing the fundamental causes of untrustworthy geometry and color predictions during volume rendering. As shown in ~\Cref{figure_12} (a-b), NeRF-S suffers from drifted sampling points near the surface (as black dashed box) and unstable, incorrect predictions before volume rendering (as black solid box), compared with vanilla NeRF. Intuitively, such uncertainty leads to foggy geometry and color errors in rendered images. Based on the above observations, we speculate that the uncertainty of surface and prediction in NeRF-S arises from the confusion of “WHERE to sample?” and “HOW to predict?”, which significantly contributes to rendering collapse when dealing with sparse inputs. Unfortunately, existing works primarily focus on regularizing rendered images after volume rendering and pay little attention to such confusion during the volume rendering process.

To address this issue, we propose the following two modules to mitigate such \textbf{confusion} (as~\Cref{figure_1}(b)):

\textbf{WHERE to sample in NeRF-S?} Due to the lack of necessary information for geometric 3D representation in NeRF-S, we introduce the prior knowledge of surface uniqueness and propose a Deformable Sampling strategy. This strategy employs a learnable offset to adaptively deform sampling positions, forcing density distribution to converge on the target accurate surface. Additionally, by analyzing the correlation between weight and offset, we design a Weight-based Mutual Information Loss to maintain rendering stability during the training phase.

\textbf{HOW to predict in NeRF-S?} To mitigate the over-fitting caused by the gap between training and testing viewpoints, we draw inspiration from consistency regularization in semi-supervised learning and propose a Semi-Supervised NeRF learning Paradigm based on pose perturbation. This paradigm adopts consistency regularization between unseen ray and perturbed rays generated by pose perturbation. To ensure local consistency, we introduce the Pixel-Patch Correspondence Loss (PPC) to alleviate unregistered pixel-wise correspondences from perturbation and improve the smoothness between adjacent rays. 

By integrating the aforementioned modules in parallel, we mitigate the uncertainty of surface and prediction, as shown in~\Cref{figure_12} (d).
 Furthermore, compared to baseline, as shown in~\Cref{figure_12} (b) and existing work based on rendered image, RegNeRF~\cite{niemeyer2022regnerf}, as shown in~\Cref{figure_12} (c), our proposed method answers the confusion of "Where" and "How" better. Its superiority is further demonstrated qualitatively and quantitatively, as detailed in~\Cref{Experiments}. In summary, we make the following specific contributions in the paper:
\begin{itemize}[leftmargin=.5cm]
    \item We demonstrate that the confusion of "WHERE to sample?" and "HOW to predict?" in NeRF-S are the main sources of rendering collapse in sparse input scenarios. Based on this insight, we propose WaH-NeRF, aiming to mitigate confusion before volume rendering.
    \item We propose a Deformable Sampling strategy and a Weight-based Mutual Information Loss to alleviate uncertainty in sample \emph{positions}. Meanwhile, we develop a Semi-Supervised NeRF Paradigm based on pose perturbation and a Pixel-Patch Correspondence Loss to mitigate uncertainty in sample \emph{predictions}.
    \item We show that WaH-NeRF achieves state-of-the-art performance compared to existing algorithms through extensive experiments, without any pre-trained models and time-consuming image warping. Our experimental results also demonstrate the effectiveness of these modules and losses in mitigating the "WHERE" and "HOW" confusion, respectively
\end{itemize}

\section{Related Work}\label{Related Work}

\subsection{Novel View Synthesis}

Novel View Synthesis (NVS) is a fundamental problem in computer vision that has been extensively researched for years \cite{debevec1996modeling,debevec1998efficient,zitnick2004high}. The purpose of NVS is generating images of unseen viewpoints by utilizing multiple images from seen viewpoints. Given multiple seen viewpoints, a straightforward idea is to first reconstruct the 3D scene and then project it onto 2D images under unseen poses~\cite{martin2018lookingood,meshry2019neural,zhou2018stereo}. For example, Penner and Zhang~\cite{penner2017soft} directly transformed the images into 3D space with the aid of initial depth images, while more recent works had favored the use of deep neural networks (DNN)~\cite{kalantari2016learning,flynn2019deepview,lombardi2019neural}. Zhou \emph{et al}.~\cite{zhou2018stereo} employed a learning framework for MultiPlane Images (MPI) representation, and~\cite{sajjadi2022scene} adopted a transformer-based encoder and decoder to learn a scalable implicit representation with attention architecture.

Recently, single-view input for NVS has gained traction as obtaining images from multiple viewpoints is not always feasible. In this scenario, explicit or implicit 3D representations are often learned from single-view input. ~\cite{tucker2020single,yoon2020novel,wiles2020synsin}. Unlike multi-viewpoint settings, other works synthesised novel views using purely image-to-image transformations without 3D representations~\cite{sun2018multi,tatarchenko2016multi}. Despite the significant progress made by NVS in both multi-view and single-view settings, the emergence of Neural Radiance Fields (NeRF)~\cite{mildenhall2021nerf} has brought new research perspectives to this field due to its simplicity and state-of-the-art performance. In contrast to most of the aforementioned NVS works that only infer discrete viewpoints, NeRF can render 360-degree continuous viewpoints without explicit representations.

\subsection{Neural Radiance Field from Sparse Inputs}

NeRF~\cite{mildenhall2021nerf} has shown a remarkable ability to produce photorealistic renderings of unseen viewpoints under the supervision of images from numerous input views. However, its performance drastically decreases when the number of input views is limited. Recent works addressed this challenge by regularizing rendered color or depth images. DSNeRF~\cite{deng2022depth} and DDNeRF~\cite{roessle2022dense} advocated the introduction of depth supervision by structure-from-motion (sfm) to compensate for the lack of information in sparse views, while DietNeRF~\cite{jain2021putting} and sinNeRF~\cite{xu2022sinnerf} adopted the consistency of semantic information to constrain the optimization process of NeRF. However, sfm struggles to function with large viewpoint gaps, and obtaining additional semantic supervision is often difficult. To address this, InfoNeRF~\cite{kim2022infonerf} minimized ray entropy for adjacent rays via KL-divergence to avoid introducing other information sources. Additionally, some studies focused on adding unseen-viewpoints-constraints to endow NeRF with prior information using geometry-smooth ~\cite{niemeyer2022regnerf} or image-warping~\cite{xu2022sinnerf} techniques. RegNeRF~\cite{niemeyer2022regnerf} proposed to regularize the color predictions at unseen viewpoints by a pre-trained flow model and demonstrated that mip-NeRF~\cite{barron2021mip} produced more satisfactory results than vanilla NeRF under sparse view settings. 

Similar to NeRF-S, other works have focused on exploring the relationship between the target viewpoints and few reference viewpoints ~\cite{yu2021pixelnerf,kulhanek2022viewformer,chibane2021stereo,wang2021ibrnet,chen2021mvsnerf}, called generalizable NeRF. Despite achieving photorealistic renderings and fast generalization capabilities~\cite{wang2022attention}, the assumption that the reference views and target view are adjacent-viewpoints is overly idealized and often requires pre-training before fine-tuning. In this work, we primarily focus on rendering a complete scene under sparse inputs.

In light of the above discussion, we utilize mip-NeRF as the baseline (see~\Cref{Method} for details) in this work. Instead of constraining results after volume rendering, our approach delves into the root cause of NeRF's confusion from sparse inputs and employs a sampling point self-alignment strategy along with a semi-supervised paradigm to overcome this challenge during volume rendering. Additionally, we follow the experimental settings of InfoNeRF ~\cite{kim2022infonerf}, rendering a complete scene without pre-training model and mainstream image-warping techniques that are time-consuming during the training phase.

\begin{figure*}
\centering
\includegraphics[width=0.8\linewidth,scale=1.00]{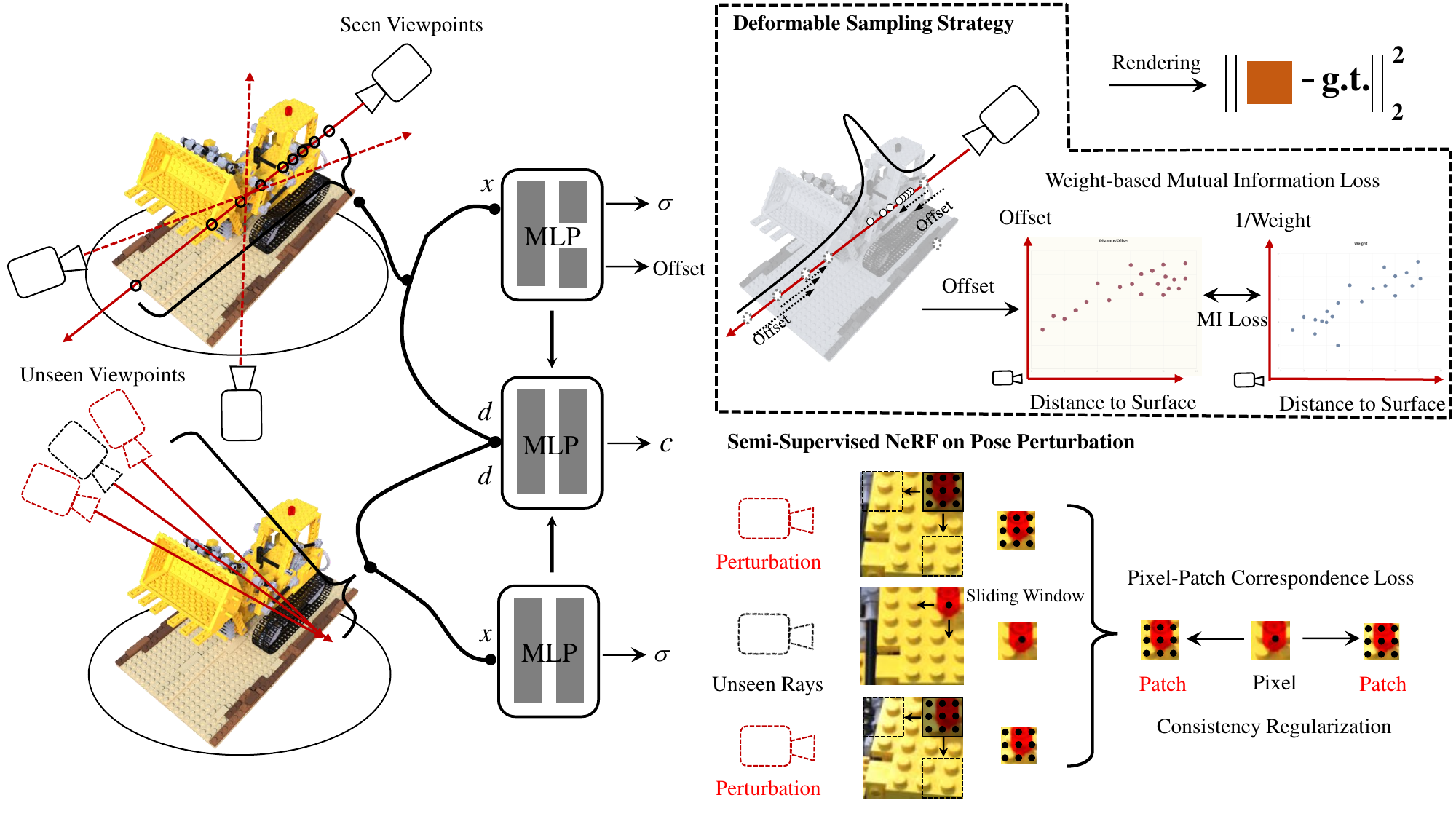}
\caption{\textit{Overall architecture of WaH-NeRF.} Firstly, for seen viewpoints, we propose a Deformable Sampling strategy and compute an offset for each sampling conical frustum, supervised by a Weight-based Mutual Information Loss. In addition, we adopt the MSE Loss with the ground truth to further constrain the training process. For unseen viewpoints, after perturbing the camera pose of these viewpoints, we use the Pixel-Patch Correspondence Loss to regulate consistency with the unseen ray for semi-supervised learning. Finally, the two modules are integrated together to mitigate the confusion of "Where" and "How".}
\label{figure_2}
\end{figure*}

\section{Method} \label{Method}
\subsection{Background}
\subsubsection{NeRF}
Neural Radiance Fields (NeRF) map the 3D location $\mathbf{x} \in \mathbb{R} ^{3} $ and viewing direction $\mathbf{d} \in \mathbb{S} ^{2} $ of the spatial point sampled on the ray to the view-invariant volume density $\sigma \in \left [ 0,\infty \right ) $ and view-dependent color $c \in \left [ 0,1 \right ]  ^{3} $ of the point using learnable Multi-Layer Perceptrons (MLPs). 
\begin{equation}
f_{\theta } \left( f_\psi\left ( \psi\left ( \mathbf{x}  \right )\right), \kappa \left ( \mathbf{d}  \right ) \right ) = \left ( \sigma, c   \right ), \label{eq1}
\end{equation}
where $f_{\theta }$ and $f_\psi$ are MLPs, the learnable mapping functions, and $\psi, \kappa$ are the positional encoding for $\mathbf{x} , \mathbf{d} $, respectively. 

\subsubsection{mip-NeRF}
Unlike NeRF's ray sampling, mip-NeRF~\cite{barron2021mip} introduces conical frustum sampling to enhance the receptive field of NeRF. Integrated positional encoding is adopted to replace $\psi, \kappa$ in conical frustums with a multivariate Gaussian. In mip-NeRF, the context of the points is taken into account during the volume rendering for each pixel, compensating for the instability when providing sparse viewpoints. Therefore, mip-NeRF serves as the baseline in our work.

\subsubsection{Volume Rendering}
Once the $c_{i}$ and $\sigma_{i}$ values for each point on the rays or cones are obtained using~\Cref{eq1}, the pixel color value $\hat{C}$ of the rendered image is approximated by an integral, which is replaced by the quadrature along the rays or cones. 
\begin{equation}
\hat{C}(r) = \sum_{i=1}^{M} (\sum_{j=1}^{i-1}\sigma_{j}\delta_{j})(1-e^{-\sigma_{i}\delta_{i}})c_{i}, \label{eq2}
\end{equation}
where $M$ is the number of sampling points, $\delta_{i}$ states the distance between the $i^{th}$ point and its adjacent point, and $r$ represents a ray or cone from an origin. 
\begin{equation}
Loss_{MSE} = \sum_{r\in R} \left \| \hat{C}(r) - C_{GT}(r)  \right \|_{2}^{2}. \label{eq3}
\end{equation}

\Cref{eq3} provides a Mean Square Error (MSE) loss to supervise the entire rendering process, where $R$ is a set containing all rays or cones and $C_{GT}(r)$ represents the ground truth color value for $\hat{C}(r)$. It is worth noting that NeRF uses a combination of coarse-sampling and fine-resampling for quadrature with non-sharing parameters, whereas mip-NeRF shares the training parameters in two stages.

\subsection{Overview}
In the following, we introduce WaH-NeRF, a novel training strategy aimed at mitigating the confusion discussed in~\Cref{Introduction} and analyzed in~\Cref{Experiments}. To explore the context of the cone, a patch-based strategy is used throughout the training process for both seen and unseen viewpoint-inputs. We then design a Deformable Sampling strategy (in~\Cref{3.3}) and Semi-Supervised NeRF based on Pose perturbation (in ~\Cref{3.4}) to address the confusion of “WHERE to sample?” and “HOW to predict?” in NeRF-S, as shown in~\Cref{figure_2}. To accurately supervise these two modules, we adopt a Weight-based Mutual Information Loss and a Pixel-Patch Correspondence Loss, which form two parallel branches during training.

\subsection{Deformable Sampling Strategy}\label{3.3}
 Recall that NeRF~\cite{mildenhall2021nerf} introduces a coarse-to-fine hierarchical sampling strategy, which achieves satisfactory results for rendering. However, this strategy is flawed when handling sparse inputs for the following reasons: 1) the difference between the training and testing viewpoints leads to overfitting with a fixed sampling strategy, and 2) the uncertainty of the surface resulting from coarse sampling makes accurate fine-resampling under sparse input challenging. In general, this sampling drift creates confusion about \textbf{"Where to sample?"}, turning the task of predicting target surface and color into an ill-posed problem for NeRF-S.

\textbf{Deformable Sampling strategy.} To determine \textbf{"Where to sample?"}, we propose a Deformable Sampling strategy inspired by deformable convolution ~\cite{dai2017deformable}. Instead of using a fixed sampling aggregation strategy~\cite{barron2022mip}, our method predicts a position-dependent and view-invariant offset for each conical frustum to adaptively aggregate sampling-points to the target surface. Based on this offset, we deform the position of the sampling conical frustum along the sampling cone before volume rendering, encouraging more frustums to approach the accurate surface. Specifically, we predict the adaptive offset using an additional MLP layer with intermediate features $F_\psi\left ( \mathbf{x}  \right )\in \mathbb{R} ^{M\times d}$ from $f_\psi$ in~\Cref{eq1}, where $d$ represents feature dimension: 
\begin{equation}
Offset = \omega_{0}  \times F_\psi\left ( \mathbf{x}  \right ) +  b_{0}, \label{eq7}
\end{equation}
where $Offset \in \mathbb{R} ^{M\times 1}$, and $\omega_{0}$ is the weight and $b_{0}$ is the bias of MLP layer. In order to prevent rendering collapse during training, both $\omega_{0}$ and $b_{0}$ are initialized to $10^{-7}$. Subsequently, the original sampled conical frustum shifts by the predicted $Offset$ along the cone for adaptive sampling, which is utilized for volume rendering as in~\Cref{eq2}. It should be noted that the Deformable Sampling strategy not applied to unseen perspectives to ensure training stability. 

In summary, Deformable Sampling strategy can be regarded as a more general sampling strategy, and vanilla NeRF \cite{mildenhall2021nerf} is a special case where $Offset$ is fixed to 0.

\textbf{Weight-based Mutual Information Loss.} However, training $\omega_{0}$ and $b_{0}$ from scratch is sub-optimal in practice since low-weight sample conical frustums struggle to offset towards the surface. To address this, we explore the correlation between the predicted \emph{Offset} and \emph{weight} ($w$) and use it as prior knowledge for additional supervision. Here, $w$ approximates the probability of the surface in NeRF ~\cite{mildenhall2021nerf}, as shown in~\Cref{eq8}. Intuitively, when the sampled conical frustums are far from the target surface, $w$ decreases and the \emph{Offset} increases. Therefore, the predicted \emph{Offset} has a negative correlation with the non-zero part of $w$.

\begin{equation}
w_{i} = (\sum_{j=1}^{i-1}\sigma_{j}\delta_{j})(1-e^{-\sigma_{i}\delta_{i}}). \label{eq8}
\end{equation}

Based on this analysis, we design a Weight-based Mutual Information Loss, a natural information-theoretic measure of the variables independence. Specifically, the mutual information of $Offset$ and $\frac{1}{w}$ is expressed as
\begin{equation}
\begin{aligned}
I\left ( \frac{1}{w+\varepsilon }; Offset\right ) &= H\left ( \frac{1}{w+\varepsilon } \right ) - H\left ( \frac{1}{w+\varepsilon }| Offset\right )\\
&= H\left ( Offset \right ) - H\left ( Offset| \frac{1}{w+\varepsilon }\right ), \label{eq9}
\end{aligned}
\end{equation}
where $\varepsilon$ is set to $5\times 10^{-4}$, and $H$ denotes entropy. The correlation between $Offset$ and $\frac{1}{w}$ increases through minimizing the Weight-based Mutual Information Loss. Inspired by ~\cite{zhao2019region}, we further express $Loss_{MI}$ by taking into account the correlation between sampling conical frustums,
\begin{equation}
\begin{aligned}
Loss_{MI} = -\sum_{r\in R} I\left ( \frac{1}{w+\varepsilon }; O\right )_{mask}, \label{eq10}
\end{aligned}
\end{equation}
where $mask$ is used to remove positions where $w$ is close to zero. The masked positions are ignored for $Loss_{MI}$, as the offset is meaningless for the positions too far from the surface. In addition, we introduce a distortion loss $Loss_{dist}$~\cite{barron2022mip} to further constrain the offset conical frustums approaching target surface. 
In summary, $L_{Offset}$ be expressed as
\begin{equation}
Loss_{Offset} = \lambda Loss_{MI} + Loss_{dist}, \label{eq12}
\end{equation}
where $\lambda$ is a hyperparameter for weight of $Loss_{MI}$, to $5\times 10^{-3}$.
\begin{figure}
\centering
\includegraphics[width=\linewidth,scale=0.3]{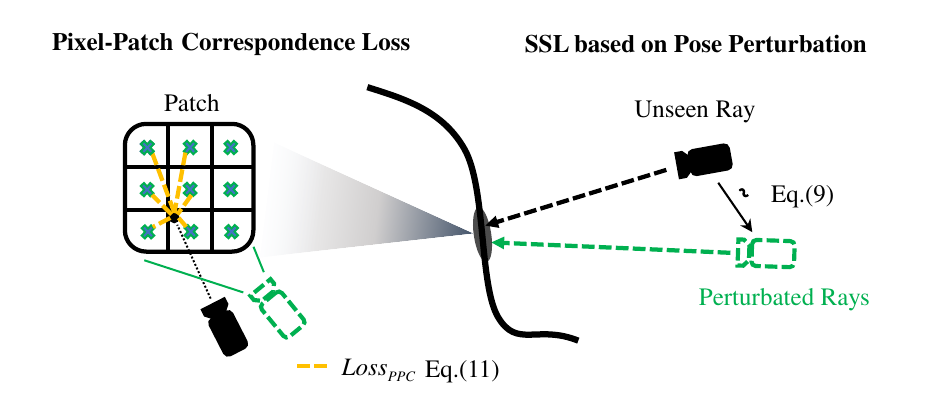}
\caption{\textit{Semi-Supervised NeRF based on pose perturbation.} We  perturb the camera position to augment limited data, and employ a Pixel-Patch Correspondence Loss as a local consistency regularization to supervise unseen ray.}
\label{figure_3}
\end{figure}
\subsection{Semi-Supervised NeRF on 
 Perturbation}\label{3.4}
Addressing the confusion of “WHERE to sample?” in NeRF-S may be helpful, but accurate sampling predictions remain a challenge under sparse views due to overfitting and insufficient information. Therefore, it is crucial to address the confusion of \textbf{“HOW to predict?”} in NeRF-S. Intuitively, introducing additional supervision, as proposed in ~\cite{xu2022sinnerf} and~\cite{niemeyer2022regnerf}, could alleviate this confusion.

In image processing, when labeled data is scarce and a vast amount of unlabeled data is available, the semi-supervised paradigm~\cite{zhu2005semi,van2020survey} can improve performance by effectively utilizing unlabeled data for supervision. 
Inspired by semi-supervised learning (SSL), researchers have treated seen viewpoints with accurate camera poses as labeled data, while considering unseen viewpoints as unlabeled data. By generating pseudo-label images of the unseen perspectives, they provide additional supervision for NeRF. However, the existing method for generating these pseudo-labels, which relies on image warping~\cite{xu2022sinnerf}, is time-consuming and limited to small viewing changes. 

\textbf{Semi-Supervised NeRF.} We propose a Semi-Supervised NeRF Paradigm based on pose perturbation to assist prediction without image warping, as shown in~\Cref{figure_3}. Thanks to NeRF's implicit representation, it can seamlessly generate continuous views. Building on this advantage, our semi-supervised paradigm employs a straightforward scheme that uses consistency regularization to explore the smoothness between the unseen ray and perturbed rays.

\textbf{Pose Perturbation.} The smooth perturbation of the NeRF for unseen rays is a non-trivial discussion. Aug-NeRF~\cite{chen2022aug} adopted perturbations into three distinct levels, including input coordinate, intermediate feature and pre-rendering output for data augmentations. However, for the semi-supervised paradigm, such local perturbations fail to reflect the smoothness of the rendering process. Instead, we propose a global perturbation method by slightly moving or rotating the camera pose and leveraging the local correlation between adjacent poses to constrain the network training. Due to the minor perturbation, there is theoretical consistency between the original unseen ray and the perturbed unseen rays. The consistency regularization not only ensures the network's robustness against discrepancies between seen and unseen viewpoints, but also improves the local smoothness of the adjacent rendered rays.

Specifically, unseen viewpoints are randomly captured on a fixed-radius sphere oriented towards the spherical render target. It is well known that any position on an arbitrary sphere can be determined by three variables: the radius $F$ controlling the sphere's size, and the azimuth angle $\varphi$ and polar angle $\Gamma $ defining the position on the sphere. Based on this prior knowledge, three levels of perturbation are performed on the camera pose, $F$, $\varphi$ and $\Gamma$, which prove to be significant in our ablation experiments. We define random perturbation as
\begin{equation}
\hat{F} = F + \tau_{F}; \\
\hat{\varphi} = \varphi + \tau_{\varphi}; \\
\hat{\Gamma} = \Gamma + \tau_{\Gamma}, \label{eq4}
\end{equation}
where $\tau_{F}$, $\tau_{\varphi}$, and $\tau_{\Gamma}$ are perturbation values. Based on the unseen-perturbed viewpoints pair, we sample the corresponding unseen-perturbed rays pair. Subsequently, the perturbed rays will be input to the NeRF-network simultaneously with the corresponding original unseen ray.

\begin{table*}
    \centering
    \caption{Mean PSNR and standard deviations of every scene on the Realistic Synthetic 360\degree dataset under three inputs.}
    \scalebox{1}{
    \setlength{\tabcolsep}{1.5mm}{
    \begin{tabular}{l|cccccccc}
        \toprule
        Method  & Lego & Chair & Drums &  Ficus & Hotdog & Materials &Mic &Ship\\
        \midrule
        NeRF, 100views   & 32.54   &33.00 &25.01 &30.13&36.18&29.62 &32.91 &28.65 \\
        \midrule
        NeRF~\cite{mildenhall2021nerf}, 3views     & $8.63_{\pm 1.21}$    &  $8.45_{\pm 0.33}$ &  $7.28_{\pm 0.51}$&  $10.23_{\pm 1.73}$&  $7.89_{\pm 0.42}$ &  $9.27_{\pm 2.10}$ &  $11.52_{\pm 1.98}$ &  $6.58_{\pm 0.75}$\\
        mip-NeRF~\cite{barron2021mip}, 3views  & $17.23_{\pm 2.84}$  & $17.31_{\pm 1.94}$  & $14.33_{\pm 0.85}$ &$18.47_{\pm 2.28}$ &$17.35_{\pm 1.64}$ & $15.51_{\pm 2.83}$ & $18.02_{\pm 2.17}$ & $15.33_{\pm 
        3.31}$\\
        \midrule
        IBRNet~\cite{wang2021ibrnet} (CVPR 2021) & $12.60_{\pm 1.53}$  & $12.33_{\pm 0.94}$  & $9.89_{\pm 1.22}$ &$15.09_{\pm 3.21}$ &$11.18_{\pm 2.48}$ & $10.47_{\pm 3.05}$ & $14.77_{\pm 1.58}$ & $13.24_{\pm 
        3.09}$\\
        pixelNeRF~\cite{yu2021pixelnerf} (CVPR 2021) & $13.22_{\pm 3.17}$  & $11.26_{\pm 2.75}$  & $11.44_{\pm 1.65}$ &$15.59_{\pm 2.99}$ &$12.41_{\pm 3.88}$ & $10.32_{\pm 4.36}$ & $15.00_{\pm 2.86}$ & $13.15_{\pm 3.26}$\\
        \midrule
        DietNeRF~\cite{jain2021putting} (ICCV 2021) & $14.98_{\pm 2.81}$ & $15.11_{\pm 2.25}$  & $12.43_{\pm 0.97}$  & $16.39_{\pm 3.10}$ & $15.21_{\pm 3.19}$ & $13.48_{\pm 2.64}$ & $14.59_{\pm 2.81}$ & $14.41_{\pm 3.22}$ \\ 
       
        InfoNeRF~\cite{kim2022infonerf} (CVPR 2022) & $16.59_{\pm 2.45}$  & $16.88_{\pm 2.04}$  & $12.16_{\pm 0.86}$ &$18.20_{\pm 2.04}$ &$17.42_{\pm 3.22}$ & $14.77_{\pm 2.98}$ & $16.73_{\pm 2.94}$ & $14.99_{\pm 3.05}$\\
        RegNeRF~\cite{niemeyer2022regnerf} (CVPR 2022) & $17.82_{\pm 2.09}$  & $17.96_{\pm 1.77}$  & $14.20_{\pm 0.91}$ &$19.01_{\pm 1.74}$ &$17.98_{\pm 3.69}$ & $15.35_{\pm 3.21}$ & $18.11_{\pm 2.03}$ & $15.82_{\pm 2.79}$\\
        WaH-NeRF (ours) & \bm{$18.78_{\pm 1.98}$}  & \bm{$18.48_{\pm 1.96}$}  & \bm{$15.39_{\pm 0.96}$} &\bm{$19.52_{\pm 1.70}$} &\bm{$19.01_{\pm 3.23}$} &\bm{$16.84_{\pm 2.45}$} & \bm{$18.95_{\pm 1.75}$} &\bm{$16.20_{\pm 2.79}$}\\
        WaH-NeRF{\dag} (ours) & \bm{$20.27_{\pm 1.90}$}  & \bm{$20.53_{\pm 1.69}$}  & \bm{$16.77_{\pm 0.87}$} &\bm{$21.83_{\pm 1.40}$} &\bm{$22.71_{\pm 2.66}$} &\bm{$18.92_{\pm 2.11}$} & \bm{$20.18_{\pm 1.68}$} &\bm{$17.91_{\pm 2.63}$}\\
        \bottomrule
    \end{tabular}}}
    \label{tab1}
\end{table*}
\textbf{Pixel-Patch Correspondence Loss.} A straightforward idea for supervision in our semi-supervised paradigm is to constrain the consistency of the rendered results between unseen ray and perturbation rays at the pixel level,
\begin{equation}
 Loss_{SSL} = \sum_{p=0}^{P} \sum_{r\in R }\left \| \hat{C}_{unseen}\left ( r \right ) - \hat{C}_{perturbation}^{p} \left ( r \right ) \right \|_{2}^{2}, \label{eq5}
\end{equation}
where $P$ indicates the amount of perturbation for an unseen ray. Despite the global similarity of the rendered images, vanilla MSE loss struggles to achieve pixel-wise (local) correspondence for consistency regularization, due to the mismatches caused by minor camera pose perturbations. 

To explore local consistency, the Pixel-Patch Correspondence Loss (PPC) is introduced to alleviate unregistered pixel-wise correspondences from perturbation and enhance patch-wise smoothness. Thanks to the patch-wise training paradigm, Pixel-Patch Correspondence loss is easily accessible. We employ the average MSE loss between each unseen ray and all rays of the corresponding patch around the perturbed rays,  

\begin{equation}
\begin{aligned}
Loss_{PPC} = \sum_{p=0}^{P} \sum_{r\in R }\sum_{i=0}^{PS\times PS} \left \| \hat{C}_{unseen}\left ( r \right ) - \right.\\
\left.\hat{C}_{perturbation\left ( i \right ) }^{p} \left ( r \right ) \right \| _{2}^{2}\times \frac{1}{PS\times PS}, \label{eq6}
\end{aligned}
\end{equation}
where $\hat{C}_{perturbation\left ( i \right ) }^{p} \left ( r \right )$ is the $i^{th}$ pixel (ray) in the perturbed patch corresponding $r$, and $PS$ is the patch size. It is worth noting that the consistency regularization not only applies to color value predictions, but also to depth predictions. Therefore, the overall Pixel-Patch Correspondence Loss is defined as the sum of the RGB and depth losses, i.e., $Loss_{PPC} = Loss_{PPC_{RGB}} + Loss_{PPC_{D}}$. This is demonstrated in the Appendix ablation experiment section.
\begin{figure}
\centering
\includegraphics[width=\linewidth,scale=0.3]{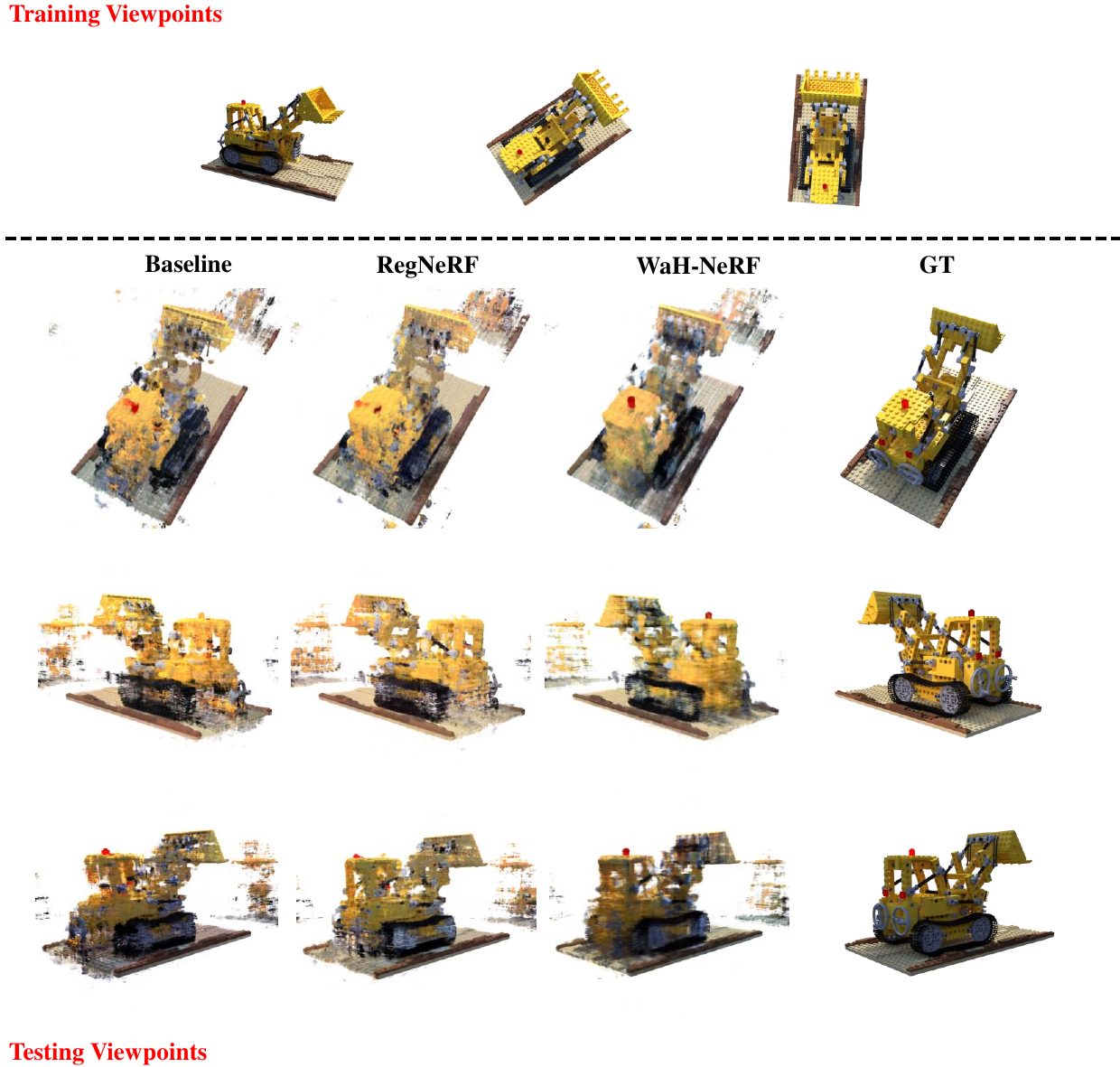}
\caption{\textit{Qualitative results from lego scene.} We choose some testing viewpoints that are significantly different from the training viewpoints under our setting.}
\label{figure_5}
\end{figure}

\begin{table}
    \centering
    \caption{Mean PSNR, SSIM and LPIPS in all scenes for quantitative evaluations and component analysis for \textbf{Realistic Synthetic 360\degree} benchmark under three inputs. DS and SSN represent Deformable Sampling strategy and Semi-Supervised NeRF Paradigm, respectively.}
    \scalebox{1}{
    \setlength{\tabcolsep}{1.5mm}{
    \begin{tabular}{l|cc|ccc}
        \toprule
        Method  & DS & SSN & PSNR $\uparrow$ &  SSIM $\uparrow$ & LPIPS $\downarrow$\\
        \midrule
        NeRF,100views  & & &31.01   & 0.947  & 0.081 \\
        \midrule
        NeRF,3views     & & &8.73   &0.445   &0.580\\
        mip-NeRF,3views  &      &  & 16.69 & 0.708 & 0.283\\
        \midrule
        IBRNet & & & 12.44   & 0.542   & 0.403 \\
        pixelNeRF &   &  &12.80 & 0.587 & 0.328\\
        DietNeRF &   &  &14.58  & 0.608 & 0.317\\
        \midrule
        InfoNeRF  &      & &15.97 & 0.647 & 0.291  \\
        RegNeRF &  &  &17.03 &0.720 &0.278\\
        \midrule
        baseline&  & &16.72 & 0.711  &0.290\\
        WaH-NeRF  & \checkmark & & 17.51 &0.768 &0.246\\
        WaH-NeRF  &  &\checkmark &17.69 & 0.773 &0.237\\ 
        WaH-NeRF (ours) &\checkmark  &\checkmark & \textbf{17.90} & \textbf{0.776} & \textbf{0.230}\\
        WaH-NeRF{\dag} (ours) &\checkmark  &\checkmark & \textbf{19.89} & \textbf{0.809} & \textbf{0.177} 
        \\
        \bottomrule
    \end{tabular}}}
    \label{tab2}
\end{table}

\subsection{Loss Function}
Given a set of $N$ seen viewpoints and their associated images, our work combines the Deformable Sampling strategy with the Semi-Supervised NeRF Paradigm to iteratively optimize the learnable parameters $\theta$ by evaluating the rendered images and calculating the loss function.
\begin{equation}
\begin{aligned}
Loss = Loss_{MSE} + \mu Loss_{Offset} +\nu Loss_{PPC} +  Loss_{smooth} . \label{eq13}
\end{aligned}
\end{equation}
As in~\cite{niemeyer2022regnerf}, $Loss_{smooth}$ is introduced to improve the smoothness of depth predictions for unseen viewpoints, which has proven effective in NeRF-S. Hyper-parameters $\mu$ and $\nu$ are used to weight the loss terms. We define $\mu = 0.5$ and $\nu = 0.5$, whose superiority is proved in the Appendix. In practice, each item in~\Cref{eq13} includes two stages of loss: coarse-sampling loss and fine-resampling loss. During training, we set the coefficient of the coarse-sampling loss to 0.1.

\section{Experiments} \label{Experiments}
\subsection{Datasets}
In order to verify the superiority of WaH-NeRF and the effectiveness of each module, we conduct comparative and ablation experiments on the Realistic Synthetic 360\degree ~\cite{mildenhall2021nerf} and LLFF~\cite{mildenhall2021nerf} datasets.\\
\textbf{Realistic Synthetic 360\degree} is a classic dataset of NeRF, which is widely used in NeRF and its derivative works. It contains 8 synthetic objects with viewpoints sampled on the upper hemisphere and coverages 360\degree. Each scene has 100 views for training and 200 views for testing, with all images at 800×800 resolution. In our task setting, due to the wide range of viewpoints, we use this dataset as the main benchmark in our work.\\
\textbf{LLFF} consists of 8 complex scenes, each containing 20-62 images
with 1008×756 resolution. Compared with Realistic Synthetic 360\degree, the real-world dataset LLFF is more complex in real-world but the viewpoints are relatively close to each other.

\subsection{Implementation and Evaluation}
\textbf{Implementation details.} We implement our code on top of the Pytorch~\cite{paszke2019pytorch} mip-NeRF \footnote{https://github.com/bebeal/mip-NeRF-pytorch}. We optimize using Adam~\cite{kingma2014adam} with an exponential learning rate decay from $10^{-3}$ to $5\times 10^{-5}$, decaying exponentially by a factor of 10 at every 2500 iterations. For the Realistic Synthetic 360\degree, we train for a total of 20,000 iterations and save the model every 5,000 iterations. We set the patch size of the seen and unseen viewpoints to be 8x8, and the batch size to 16, which means that both input 1024 rays, $\left | R \right | = 1024$. Following the experimental setup of InfoNeRF ~\cite{kim2022infonerf}, under $N=3$ setting, we randomly select three viewpoints for training, and all comparison experiments use the same three viewpoints to ensure fairness. Our experiments are conducted with a GeForce RTX 3090 GPU.\\
\textbf{Metrics.} We employ the standard image quality metrics, including Peak Signal to-Noise Ratio (PSNR) and Structural SIMilarity (SSIM) ~\cite{wang2004image}, to evaluate rendering quality for novel viewpoints. Additionally, we introduce learned perceptual image patch similarity (LPIPS) ~\cite{zhang2018unreasonable} as a perceptual metric. For all metrics, we calculate the mean and standard deviation for comparative experiments.\\
\textbf{Baseline.} We use the Pytorch reimplementation of mip-NeRF~\cite{barron2021mip} as the baseline network, which has been shown to outperform vanilla NeRF in our experiments. We further integrate a patch-based training strategy into the mip-NeRF as the baseline. In the comparative experiments, we evaluate our method against vanilla NeRF ~\cite{mildenhall2021nerf} and mip-NeRF~\cite{barron2021mip} under sparse inputs, and the state-of-the-art models under our setting, including InfoNeRF~\cite{kim2022infonerf}, DietNeRF~\cite{jain2021putting}, DSNeRF~\cite{deng2022depth}\footnote{Due to the use of sfm and COLMAP, DSNeRF is only used for comparative experiments on the LLFF.} and RegNeRF~\cite{niemeyer2022regnerf}. Additionally, we compare our approach against some recent works, namely IBRNet ~\cite{wang2021ibrnet} and pixelNeRF~\cite{yu2021pixelnerf} on the adjacent-viewpoints setting, which has been shown to outperform traditional NVS methods. For a detailed description about these, see~\Cref{Related Work} and Appendix.

\begin{figure}
\centering
\includegraphics[width=\linewidth,scale=0.3]{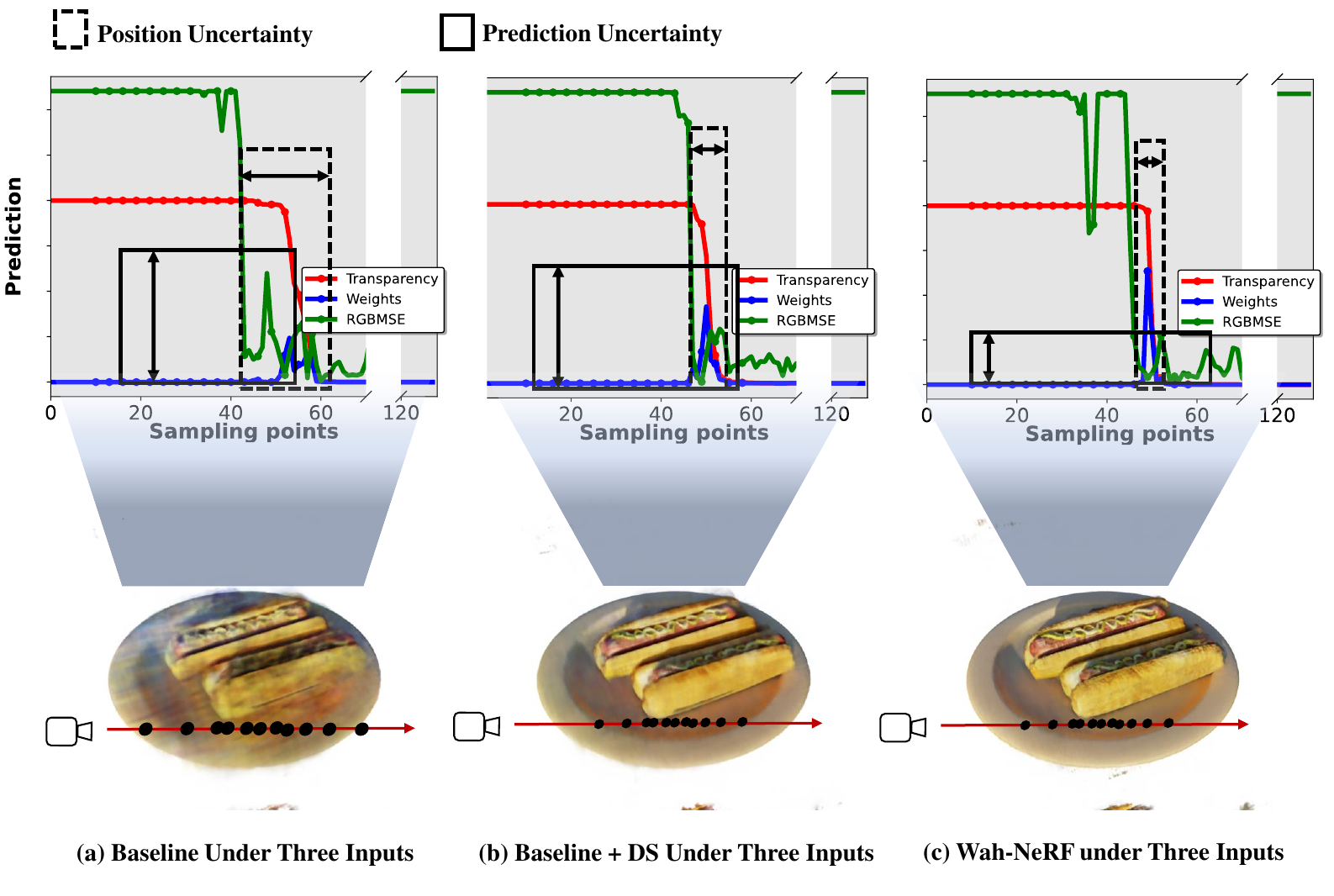}
\caption{Qualitatively demonstrate that the proposed module can effectively alleviate uncertainty of surface position and prediction at hotdog scene.}
\label{figure_9}
\end{figure}

\subsection{Experiment and Analysis}
\textbf{Quantitative evaluations results.} ~\Cref{tab1} and ~\Cref{tab2} present the performance comparison between our proposed WaH-NeRF and the leading methods on the Realistic Synthetic 360\degree dataset. It can be observed that WaH-NeRF achieves state-of-the-art performance under three inputs ($N=3$) for each scene. Specifically, summarizing from~\Cref{tab2}, WaH-NeRF surpasses the previous state-of-the-art method (i.e., RegNeRF) by \textbf{0.87} (PSNR), \textbf{0.046} (SSIM) and \textbf{0.048} (LPIPS). Additionally, compared with the vanilla NeRF and mip-NeRF, our WaH-NeRF achieves significant improvement, i.e.,\textbf{1.21} (PSNR), \textbf{0.068} (SSIM) and \textbf{0.053} (LPIPS) for mip-NeRF under our settings. Moreover, we also conduct experiments under different numbers of inputs in the following section. These results demonstrate the effectiveness and superiority of our proposed WaH-NeRF over the existing state-of-the-art methods. Here, WaH-NeRF{\dag} represents the WaH-NeRF's rendering performance after masking the background. This shows that WaH-NeRF can achieve satisfactory performance for the foreground object with only three inputs.\\
\textbf{Qualitative evaluation results.} ~\Cref{figure_5} reports qualitative results rendered by WaH-NeRF on the lego scene, providing additional evidence of its effectiveness. Intuitively, compared with the state-of-the-art RegNeRF \cite{niemeyer2022regnerf} and the baseline, WaH-NeRF shows a clear advantage in mitigating confusion and achieving better rendering results for surface and color. More quantitative results in other scenes can be found in the Appendix. \\
\textbf{Component analysis.} Our proposed WaH-NeRF consists of two key components: the Deformable Sampling strategy and the Semi-Supervised NeRF on pose perturbation. We validate the effectiveness of each component and present the results in ~\Cref{tab2}. It can be summarized that the Semi-Supervised NeRF Paradigm plays the important role in the rendering performance improvement while the Deformable Sampling strategy is indispensable. By combining both Semi-Supervised NeRF Paradigm and Deformable Sampling strategy, WaH-NeRF achieves a substantial improvement of \textbf{1.18} (PSNR), \textbf{0.065} (SSIM), and \textbf{0.060} (LPIPS) over the baseline model. 

To further explain the benefits of the proposed components during volume rendering, as shown in~\Cref{figure_9}, we qualitatively demonstrate how the Deformable Sampling strategy and the Semi-Supervised NeRF Paradigm help to alleviate the confusion of “WHERE to sample?” and “HOW to predict?" in NeRF-S. Additional analyses and quantitative results are available in the Appendix.
\begin{table}
    \caption{Mean PSNRs, SSIM and LPIPS in all scenes for loss function ablation experiment on \textbf{Realistic Synthetic 360\degree} benchmark from three inputs.}
    \scalebox{1}{
    \setlength{\tabcolsep}{3mm}{
    \centering{
    \begin{tabular}{lccccc}
        \toprule
        Method   & PSNR $\uparrow$ &  SSIM $\uparrow$ & LPIPS $\downarrow$\\
        \midrule
        w/o $L_{PPC}$  &17.73   & 0.773   &0.239 \\
        w/o $L_{Offset}$     &17.30   &0.742   &0.265\\
        \midrule
        WaHNeRF (ours)  & \textbf{17.90} & \textbf{0.776} & \textbf{0.230} \\
        \bottomrule
    \end{tabular}}}}
    \label{tab4}
\end{table}
\\
\textbf{Loss functions analysis.} $L_{Offset}$ and $L_{PPC}$ are important components in our loss function. In~\Cref{tab4}, the significant contribution of both losses to the network is evident, with $L_{Offset}$ being particularly noteworthy. This could be attributed to the difficulty of training the $Offset$ from scratch without the aid of $L_{Offset}$.
\\
\begin{figure}
\centering
\includegraphics[width=0.9\linewidth,scale=0.3]{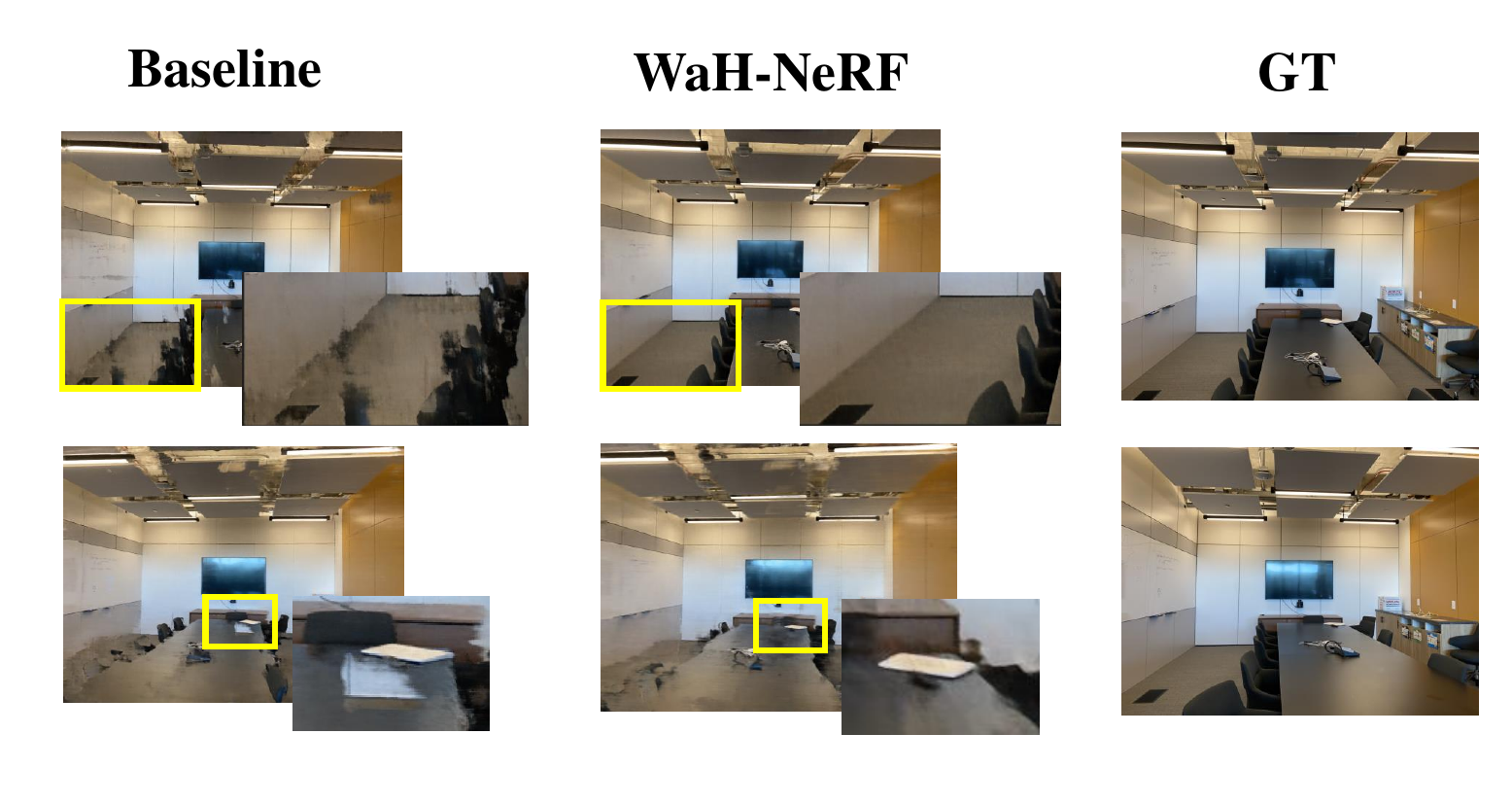}
\caption{Some qualitative results of the \textbf{baseline} and \textbf{WaH-NeRF} under LLFF dataset (room scene) in the 3-Inputs setting.}
\label{figure_8}
\end{figure}
\textbf{Quantitative and qualitative results for LLFF.} \Cref{tab:3} shows the quantitative comparison between the previous state-of-the-art methods and WaH-NeRF on the LLFF dataset, where FT representing Fine-Tuning on per scene~\cite{niemeyer2022regnerf}. Despite the LLFF dataset being more challenging due to its larger scene size and the presence of only nearby viewpoints, WaH-NeRF also attains state-of-the-art results. This demonstrates the superiority of method regardless of the task scenario. \Cref{figure_8} showcase some qualitative results rendered by WaH-NeRF on the room scene, providing further evidence of its effectiveness. Additional quantitative results for other LLFF scenes are available in the Appendix.

\begin{table}
    \centering
    \caption{Mean PSNRs, SSIM and LPIPS in all scenes for quantitative evaluations for \textbf{LLFF} benchmark.}
    \scalebox{1}{
    \setlength{\tabcolsep}{1.5mm}{
    \begin{tabular}{l|c|c|ccc}
        \toprule
        Method  &$N$ & Setting & PSNR $\uparrow$ &  SSIM $\uparrow$ & LPIPS $\downarrow$\\
        \midrule
        NeRF &100&  &26.50   & 0.811  & 0.250 \\
        \midrule
        pixelNeRF  &3 & \multirow{2}{*}{\shortstack{Trained on \\ DTU and FT}} &16.17   & 0.438  & 0.512 \\
        MVSNeRF   &3 &  &17.88   & 0.584  & 0.327 \\
        \midrule
        mip-NeRF &3 & \multirow{6}{*}{\shortstack{Optimized \\ per Scene}}   & 14.62 & 0.351 & 0.495\\
        DietNeRF &3 &   & 14.94 & 0.370 & 0.496\\
        InfoNeRF &3 &   &14.37 & 0.349 & 0.457  \\
        DSNeRF &3 &   &18.51 & 0.558 & 0.338  \\
        RegNeRF &3 &  &19.08  &  0.587  & 0.336 \\
        WaH-NeRF &3 & & \textbf{19.23} & \textbf{0.587} & \textbf{0.312} \\
        \bottomrule
    \end{tabular}}}
    \label{tab:3}
\end{table}

\section{Conclusion}
In this paper, we propose a novel training paradigm for Neural Radiance Fields from sparse inputs, termd as WaH-NeRF. We observed that NeRF's performance significantly declines as the number of input views decreases. Our analysis attribute this phenomenon to the confusion of "WHERE to sample?" and "HOW to predict?" in NeRF-S. To address this confusion, we introduce a Deformable Sampling strategy and a Semi-Supervised NeRF learning Paradigm based on pose perturbation, leveraging the Pixel-Patch Correspondence Loss and the Wight-based Mutual Information Loss in the training process. Extensive experiments on two NeRF benchmarks demonstrate that our proposed WaH-NeRF achieves state-of-the-art performance under sparse input settings, as supported by both quantitative and qualitative results. We believe that the confusion analysis for NeRF-S and the semi-supervised NeRF based on consistency regularization will inspire future research in this field.


\bibliographystyle{ACM-Reference-Format}
\balance
\bibliography{submit}


\end{document}